\documentclass[10pt,twocolumn,letterpaper]{article}

\usepackage{cvpr}              % To produce the CAMERA-READY version
\usepackage[table,xcdraw]{xcolor}  
\usepackage{colortbl}              
\usepackage{soul}                  
\usepackage{booktabs}              

\definecolor{colorfirst}{rgb}{.866,.945,0.831}
\definecolor{colorsecond}{rgb}{1,0.98,0.83}
\definecolor{colorthird}{rgb}{0.76,0.87,0.92}

\newcommand{\cellfirst}[1]{\cellcolor{colorfirst}{#1}}
\newcommand{\cellsecond}[1]{\cellcolor{colorsecond}{#1}}

\DeclareRobustCommand{\textfirst}[1]{\sethlcolor{colorfirst}\hl{#1}}
\DeclareRobustCommand{\secondtext}[1]{\sethlcolor{colorsecond}\hl{#1}}

\usepackage{booktabs}      
\usepackage{tabularx}      
\usepackage{array}         
\usepackage{pifont}       
\newcolumntype{Y}{>{\raggedright\arraybackslash}X}

\newcommand{\greencheck}{\textcolor{green!60!black}{\ding{51}}}
\newcommand{\redx}{\textcolor{red!70!black}{\ding{55}}}

\definecolor{colorblue}{rgb}{0.2,0.45,0.8}  
\definecolor{colororange}{rgb}{1.0,0.55,0.15} 

\usepackage{algorithm}
\usepackage{algpseudocode}

\usepackage[accsupp]{axessibility} % Improves PDF readability for those with disabilities.
\usepackage[misc]{ifsym}
\definecolor{cvprblue}{rgb}{0.21,0.49,0.74}
\usepackage[pagebackref,breaklinks,colorlinks,allcolors=cvprblue]{hyperref}

\def\paperID{7529} % *** Enter the Paper ID here
\def\confName{CVPR}
\def\confYear{2026}

\title{Mocap-2-to-3: Multi-view Lifting for Monocular Motion Recovery with 2D Pretraining}

\author{
Zhumei Wang$^{1,2}$\thanks{Equal contribution.} \quad
Zechen Hu$^{3}$\footnotemark[1] \quad
Ruoxi Guo$^{4}$ \quad
Huaijin Pi$^{5}$ \quad
Ziyong Feng$^{3}$ \quad  \\
Liang Zhang$^{6}$ \textsuperscript{\Letter} \quad
Mingtao Pei$^{1}$ \quad
Siyuan Huang$^{2}$ \textsuperscript{\Letter} \\
{\small$^{1}$Beijing Institute of Technology  \quad
$^{2}$State Key Laboratory of General Artificial Intelligence, BIGAI} \\
{\small$^{3}$Deep Glint  \quad
$^{4}$Zhejiang University  \quad
$^{5}$The University of Hong Kong  \quad
$^{6}$Shandong Agricultural University }\\
{\small * Equal contribution \qquad \textsuperscript{\Letter} Corresponding authors \qquad Project page: \url{https://wangzhumei.github.io/mocap-2-to-3/}}
}
\begin{document}
% \maketitle
\twocolumn[\maketitle\vspace{-12mm}% \vspace{-10mm}
\begin{center}
    \centering
    \includegraphics[width=\linewidth]{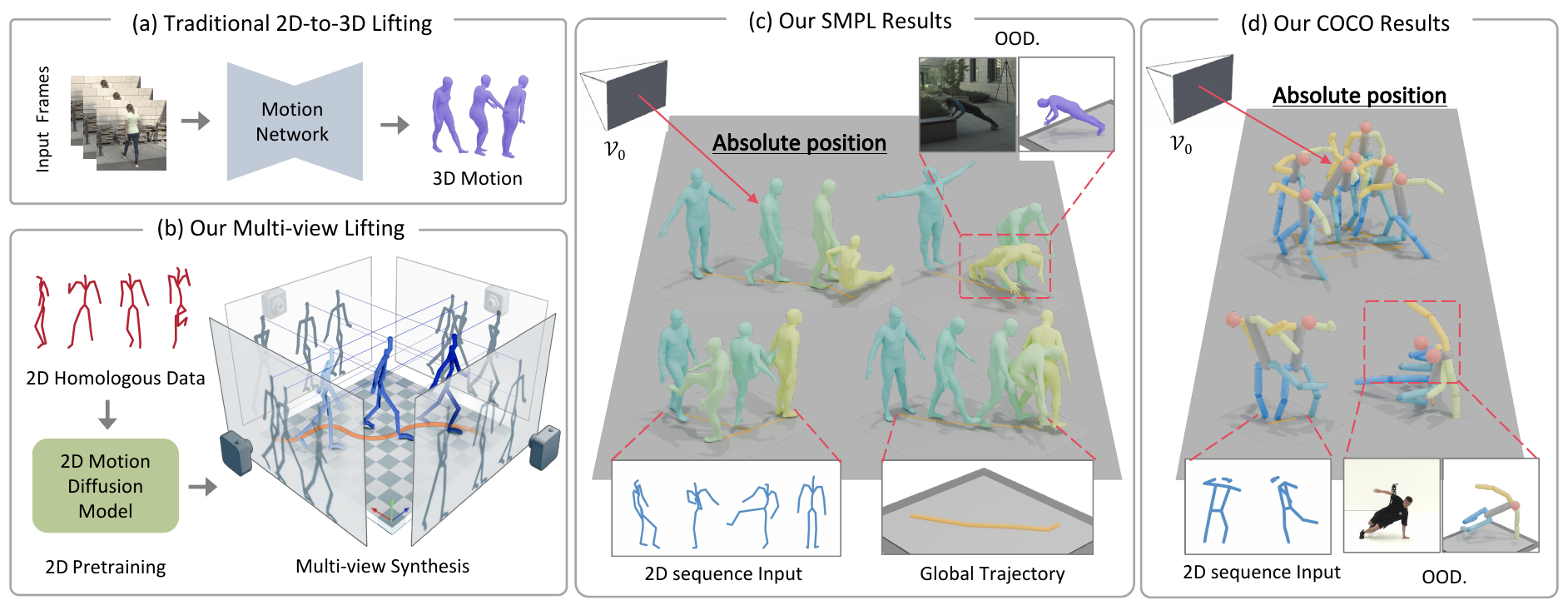}
    \vspace{-8mm}
    \captionof{figure} {(a) Traditional framework for direct 3D motion regression. (b) \textbf{Mocap-2-to-3}: our multi-view lifting framework from monocular input which leverages 2D pretraining to enhance 3D motion capture. (c) The model outputs SMPL-format global motions with \textbf{absolute position} from monocular 2D pose input while maintaining \textbf{out-of-distribution} generalization capability. (d) Our model also supports outputs in the COCO-format keypoint.
    }
    \label{fig:teaser}
    \vspace{-2mm} %1
\end{center}
\bigbreak]
\begin{abstract}
\vspace{-6mm}

% \textcolor{red}{Human motion recovery requires interaction with the physical world, demanding not only accurate skeletal structures but also metrically precise positions. 
%   However, recovering absolute human motion from monocular inputs remains challenging due to two factors: (1) reliance on 3D training data collected in constrained environments, which limits out-of-distribution generalization, and (2) the difficulty of estimating metric-scale poses from monocular observations.
%   In this paper, we discuss how to improve 3D motion accuracy solely from 2D data, and how to obtain metric precision from monocular input. % while mitigating deviations from the physical world.
%   We propose a novel framework \textbf{Mocap-2-to-3}, designed to lift monocular inputs into consistent multi-view motion representations.}
%   To leverage abundant 2D data, we decompose complex 3D motion into multi-view syntheses: a single-view diffusion model is first pretrained on extensive 2D datasets, followed by multi-view fine-tuning on public 3D data, enabling view-consistent motion generation while transferring action priors and diversity from 2D supervision.  
  Human motion recovery for real-world interaction demands both precise action details and metric-scale trajectories.
  Recovering absolute human pose from monocular input presents a viable solution, but faces two main challenges: (1) models' reliance on 3D training data from constrained environments limits their out-of-distribution generalization% and fails to leverage the diversity of 2D data
  ; and (2) the inherent difficulty of estimating metric-scale poses from monocular observations.
  % In this paper, we propose Mocap-2-to-3, a novel framework based on a virtual multi-view motion representation that leverages abundant 2D data to enhance 3D motion recovery, including absolute positioning.
  % In this paper, we propose Mocap-2-to-3, a novel framework for recovering absolute pose from monocular input and enhancing 3D motion recovery by leveraging abundant 2D data.
  This paper introduces \textbf{Mocap-2-to-3}, a novel framework that differs from prior HMR methods by recovering absolute poses from monocular input and leveraging abundant 2D data to enhance 3D motion recovery.
  To effectively utilize the action \textbf{priors and diversity} in large-scale 2D datasets, we reformulate 3D motion as a multi-view synthesis process and divide the training into two stages: a single-view diffusion model is first pre-trained on extensive 2D data, followed by multi-view fine-tuning on 3D data, thus achieving a combination of strong priors and geometric constraints.
  Furthermore, to recover absolute poses, we introduce a novel human motion representation that decouples the learning of local pose and global movements, while encoding ground geometric priors to accelerate convergence, thereby yielding more precise positioning in the physical world.
  Experiments on in-the-wild benchmarks show that our method outperforms state-of-the-art approaches in both camera-space motion realism and world-grounded human positioning, while exhibiting strong generalization capability.
  % Code will be publicly available.
  % Our code will be made publicly available.
  
\end{abstract}

\vspace{-6mm}
\section{Introduction}
\label{sec:intro}
\vspace{-1mm}

Markerless motion capture for downstream tasks involving interaction with the physical world requires motion recovery in absolute coordinates, a requirement that our work addresses by reconstructing \textbf{absolute positions} in world coordinates from monocular input.
Such markerless mocap enables a wide range of applications, including gaming, sports analysis, multi-person interactions, and embodied intelligence. Compared to multi-view systems, monocular reconstruction uses less hardware, imposes fewer constraints, and is more practical for downstream tasks~\cite{shuai2023reconstructing}.

Current state-of-the-art methods \cite{shin2024wham,shen2024world,zhu2023motionbert,sun2021monocular,sun2022putting,sun2024aios} heavily rely on precise 3D motion capture data \cite{mahmood2019amass,black2023bedlam,ionescu2013human3,patel2021agora} for training, which is costly and requires specialized equipment and controlled environments, limiting accessibility for many research institutions. Moreover, the complex procedures hinder timely acquisition of out-of-distribution 3D data for downstream tasks. Since interaction with the physical world demands higher fidelity in motion details, models often require fine-tuning on domain-specific or homogeneous data to ensure accuracy. Unlike 3D data, 2D data is more accessible, easily obtained from internet videos with diverse real-world actions~\cite{pi2024motion} or through annotated or estimated 2D skeletons in specific scenarios~\cite{zhu2023motionbert,xu2022vitpose}.

While monocular motion estimation has shown strong performance in academic settings, prior methods~\cite{shin2024wham,shen2024world} recover only relative positions in the world coordinate system, typically via alignment with the ground truth’s initial frame—limiting practical deployment. In contrast, absolute positioning supports broader applications by requiring environmental awareness and spatial reasoning to infer complex interactions, including human-human, human-object, and human-scene relations. Motivated by the need for accurate interaction with the physical world in markerless motion capture, we focus on recovering global motions with absolute depth from monocular inputs(\cref{fig:teaser}(c)).

% To address this limitation, we propose Mocap-2-to-3, a diffusion-based pipeline that leverages 2D data to enhance 3D motion capture. Inspired by~\cite{pi2024motion}, we reformulate 3D motion modeling as a multi-view reconstruction task (using one real primary view and synthesized auxiliary views) and adopt a two-stage framework: (1) single-view pre-training and (2) view-consistent multi-view fine-tuning. This design confines the limitations of 3D data (e.g., priors and diversity) to the pre-training phase, while utilizing homologous 2D data to improve generalization. In Stage two, we initialize with pre-trained weights, add a View Attention Layer, and fine-tune with 3D supervision to enforce multi-view consistency. At inference, the diffusion model generates novel-view 2D poses from input 2D sequences and lifts them to 3D motion.

We propose Mocap-2-to-3, a diffusion-based model that leverages 2D data to enhance 3D motion capture. As shown in \cref{fig:teaser}(a), unlike traditional 2D-to-3D lifting methods that directly regress 3D motion from monocular input, we draw inspiration from~\cite{pi2024motion} and propose a multi-view lifting framework that synthesizes 3D motion from monocular input (Fig.~\ref{fig:teaser}(b)). This framework leverages the diversity of 2D data to overcome the poor generalization caused by training solely on limited 3D data. Our model is trained in two stages: (1) a single-view pretraining, which enriches the model’s \textbf{priors} through homologous 2D data, improving its generalization to novel scenarios and reducing motion errors in the physical world; and (2) a view-consistent multi-view fine-tuning, where a View Attention Layer is inserted to enforce multi-view consistency during the generation of other-view 2D poses from an input 2D sequence.
Within this framework, we aim to obtain \textbf{metric-scale} human poses rather than merely recovering aligned global trajectories. Given calibrated camera poses, we estimate novel-view global motion from a monocular input to synthesize 3D motion with absolute positioning. 
Direct per-view global motion generation is suboptimal, however, as loss dominance by position impedes learning of subtle variations. 
To address this, we introduce a novel motion representation that decouples local pose and global movement for independent learning, enabling the network to capture more fine-grained motion details.
Nevertheless, monocular 2D-to-3D lifting remains inherently ill-posed, as depth (Z-axis) cannot be directly inferred from 2D inputs and requires additional geometric constraints or priors. 
As a result, learning view-consistent global movement converges slowly. To overcome this, we encode the camera pose into explicit ground-plane constraints, allowing the network to learn geometric priors through cross-attention and thereby accelerate convergence. Through this process, we obtain accurate absolute human poses in the physical world.

Our framework supports lifting any 2D pose format (e.g., SMPL~\cite{huang2022capturing}, COCO~\cite{lin2014microsoft}, H36M~\cite{ionescu2013human3}) to 3D by retraining on the desired format(\cref{fig:teaser}(c)(d)). This work focuses on enhancing the \textbf{2D-to-3D lifting process}. Errors from inaccurate 2D detectors, caused by significant deviations from ground truth are beyond our scope, as we do not process raw images or apply secondary corrections. By decoupling image input from 3D motion estimation, we use limited 3D data to synthesize large-scale virtual training samples, improving generalization.

Our main contributions are as follows: (1) We propose a multi-view lifting framework from monocular input that leverages 2D pretraining to learn strong motion priors and is fine-tuned on limited 3D data to enable view-consistent generation, effectively lifting 2D knowledge to enhance 3D motion reconstruction. (2) We propose a novel human motion representation that separates local motion from global movement, enabling accurate absolute pose recovery while preserving fine-grained motion details. (3) We compute the ground-plane equation from camera poses and encode it into the network, thereby explicitly constraining human positions in the physical space and accelerating the convergence of multi-view global trajectory learning. (4) Extensive experiments demonstrate the effectiveness and generalization of our method, outperforming prior approaches in both motion accuracy and global positioning, even without scene-specific data.

% We incorporate physical-world constraints to correct human positioning and encode camera parameters as ground-plane representations, thereby accelerating the convergence of multi-view global trajectory learning.
\section{Related Works}
\label{sec:related}

\begin{table}[t]
    \centering
    \resizebox{1.0\columnwidth}{!}{%
    \begin{tabular}{l|cccc}
        \toprule
        \textbf{Method} & \textbf{Input} & \textbf{Camera Parameters} & \textbf{2D Data Leverage} & \textbf{Metric-scale Traj.} \\
        \midrule
        SMPLify~\cite{bogo2016keep} & Images & Optimization & -- & \greencheck \\
        MotionBERT~\cite{zhu2023motionbert} & 2D Pose & -- & \greencheck & \redx \\
        SA-HMR~\cite{shen2023learning} & Images, Scene Scans & Calibration & \redx  & \greencheck \\
        Ray3D~\cite{zhan2022ray3d} & 2D Pose & Calibration & \redx  & \greencheck \\
        WHAM~\cite{shin2024wham} & Images & Estimate & \redx  & \redx \\
        GVHMR~\cite{shen2024world} & Images, 2D Pose & Estimate & \redx  & \redx \\
        TRAM~\cite{wang2024tram} & Images & GT K + SLAM & \redx  & \greencheck \\
        MetricHMR~\cite{zhang2025metrichmr} & Images & GT K + SLAM & \redx  & \greencheck \\
        MVLift~\cite{li2025lifting} & 2D Pose & Specified (Virtual) & \greencheck  & \redx \\
        \midrule
        \textbf{Mocap-2-to-3 (Ours)} & 2D Pose & Calibration & \greencheck & \greencheck \\
        \bottomrule
    \end{tabular}%
    }
    \vspace{-2mm}
    \caption{\textbf{Comparison with related methods.} 
    Unlike methods limited to canonical/root-aligned trajectories, Mocap2-to-3 recovers metric-scale trajectories from monocular 2D input and can further leverage 2D data to enhance 3D results.}
    \label{tab:related_comp}
    \vspace{-6mm} %4
\end{table}

\subsection{Monocular human motion recovery}
% Early methods like HMR~\cite{kanazawa2018end} pioneered end-to-end SMPL~\cite{15tog_smpl} regression from images. Later works improved accuracy via optimization (SPIN~\cite{kolotouros2019learning}) and temporal modeling (VIBE~\cite{kocabas2020vibe}, HMMR~\cite{kanazawa2019learning}, DSD~\cite{xu2020deep}). Recent methods enhanced robustness to occlusion (PARE~\cite{kocabas2021pare}) and incorporated global cues (CLIFF~\cite{li2022cliff}). 
% Global recovery methods aim to decouple human and camera motion: WHAM~\cite{shin2024wham} predicts poses autoregressively but suffers drift; GVHMR~\cite{shen2024world} aligns motion to gravity but needs initial alignment; SLAHMR~\cite{ye2023decoupling} and PACE~\cite{kocabas2024pace} integrate SLAM and priors but are computationally expensive. ROMP~\cite{sun2021monocular} and PromptHMR~\cite{wang2025prompthmr} enables efficient multi-person recovery. HumanMM~\cite{zhang2025humanmm} requires multi-camera video capture. 
% Thus, monocular metric pose estimation is challenging, motivating 2D-driven method.

Early monocular methods such as HMR~\cite{kanazawa2018end} pioneered end-to-end regression of SMPL~\cite{15tog_smpl} parameters directly from images. Subsequent approaches improved accuracy through optimization-in-the-loop refinement (SPIN~\cite{kolotouros2019learning}) and temporal modeling on videos (HMMR~\cite{kanazawa2019learning}, VIBE~\cite{kocabas2020vibe}, DSD~\cite{xu2020deep}). Later methods enhanced robustness to occlusion (PARE~\cite{kocabas2021pare}) and incorporated global camera cues (CLIFF~\cite{li2022cliff}).
To move beyond camera space, recent “world-grounded” methods decouple human and camera motion. WHAM~\cite{shin2024wham} reconstructs temporally coherent motion trajectories by integrating video-based cues, while GVHMR~\cite{shen2024world} introduces a gravity-view coordinate system to stabilize long-term orientation. SLAHMR~\cite{ye2023decoupling} and PACE~\cite{kocabas2024pace} combine SLAM and learned motion priors for joint optimization of camera and human motion, albeit with heavy computational cost. ROMP~\cite{sun2021monocular} and PromptHMR~\cite{wang2025prompthmr} achieve efficient multi-person estimation, and HumanMM~\cite{zhang2025humanmm} leverages multi-camera data for large-scale human motion modeling.

While these approaches successfully recover global trajectories (often from video input), they generally lack metrically accurate absolute positioning in the physical world. By contrast, our goal is to retain high-fidelity action recovery and estimate absolute positions with metric scale, which is crucial for downstream tasks requiring precise interaction with the real world.

%-------------------------------------------------------------------------
\subsection{Monocular 3D absolute pose estimation}

% Several methods focus on recovering the absolute human pose in a world coordinate system from monocular input. For instance, SMPLify~\cite{bogo2016keep} obtains camera pose and human motion via optimization. Ray3D~\cite{zhan2022ray3d} maps 2D keypoints into 3D ray space using geometric constraints for accurate, robust localization. SA-HMR~\cite{shen2023learning} estimates absolute mesh positions from a single image by leveraging a pre-scanned scene, reducing ambiguities. In contrast, TRAM~\cite{wang2024tram} and MetricHMR~\cite{zhang2025metrichmr} rely on SLAM to estimate a moving camera pose for absolute motion recovery. However, such estimated poses are prone to bias, leading to cumulative errors and unreliable positional estimates for interaction. 
% Similar to \cite{zhan2022ray3d,shen2023learning}, which recover absolute poses using calibrated fixed cameras, our framework operates under comparable settings and estimates metric-scale poses from calibration.

Several methods aim to recover absolute human poses in a world coordinate system from monocular input. SMPLify~\cite{bogo2016keep} jointly optimizes camera and body parameters, while Ray3D~\cite{zhan2022ray3d} projects 2D keypoints into 3D ray space under geometric constraints to achieve accurate localization. SA-HMR~\cite{shen2023learning} infers absolute mesh positions from a single image by leveraging a pre-scanned scene to resolve scale ambiguities. In contrast, TRAM~\cite{wang2024tram} and MetricHMR~\cite{zhang2025metrichmr} integrate SLAM-based camera pose estimation for absolute motion recovery, but such estimated poses often introduce bias and accumulate drift, leading to unreliable positional accuracy for interaction.

Similar to \cite{zhan2022ray3d,shen2023learning}, which recover absolute poses using calibrated fixed cameras, our framework operates under comparable conditions. 
By leveraging calibrated camera poses, we minimize system bias and achieve more accurate metric-scale estimations, aligning with our objective of recovering physically grounded, metrically precise poses that support downstream interaction tasks.
Unlike previous methods, we propose a novel multi-view 2D-to-3D lifting framework that further decouples coordinates to recover precise absolute poses and leverages 2D pretraining to learn richer motion priors, enhancing generalization.
% Unlike SA-HMR~\cite{shen2023learning}, our method does not rely on additional scene scans as input, thereby avoiding extra costs and simplifying real-world deployment.

%-------------------------------------------------------------------------
\subsection{2D-driven Motion Generation and Recovery}

% Some works explore motion generation using diffusion models. MDM~\cite{2023Human} introduces a diffusion-based framework that captures complex dynamics via denoising. MAS~\cite{kapon2024mas} applies 2D diffusion in a multi-view setup for high-fidelity 3D motion with improved spatial consistency. Motion-2-to-3~\cite{pi2024motion} incorporates large-scale internet videos during pre-training to boost generation quality. We draw inspiration from this pre-training strategy to enhance motion capture.

% MotionBERT~\cite{zhu2023motionbert} and ElePose~\cite{wandt2022elepose} lift 2D poses to 3D, but often lack reliable global trajectories. 
% While MVLift~\cite{yin2024lifting} shows that global motion can be recovered from 2D-only training, the motion quality cannot match that of using 3D data. Our framework preserves 3D constraints while leveraging 2D data to increase diversity, resulting in improved performance and generalization.

Recent studies have explored leveraging 2D data for 3D human motion generation and recovery. MotionBERT~\cite{zhu2023motionbert} and ElePose~\cite{wandt2022elepose} lift 2D poses to 3D but often lack reliable global trajectories. Diffusion-based approaches, such as MDM~\cite{2023Human}, model temporal dynamics through iterative denoising, while MAS~\cite{kapon2024mas} extends 2D diffusion into a multi-view setting to enhance spatial consistency. Motion-2-to-3~\cite{pi2024motion} utilizes large-scale internet videos during pretraining to improve motion diversity and realism, a strategy that inspires our framework design. Similarly, MVLift~\cite{li2025lifting} demonstrates that global motion can be recovered using 2D-only training; however, the motion quality still lags behind 3D-supervised methods due to the inherent advantages of 3D data—accurate absolute positioning, coordinated dynamics, and consistent skeletal proportions. Building upon these insights, our framework integrates structured 3D data with diverse 2D data, achieving both higher performance and stronger generalization.

%-------------------------------------------------------------------------
In summary, unlike the aforementioned approaches, our method lifts monocular 2D poses to 3D motion, not only leveraging 2D data for pretraining to improve adaptability to out-of-distribution scenarios, but also recovering metrically accurate poses in the physical world. detailed comparison is provided in \cref{tab:related_comp}.
\section{Method}

\begin{figure*}[ht]
\centering%
\vspace{-5mm}
\includegraphics[width=\linewidth]{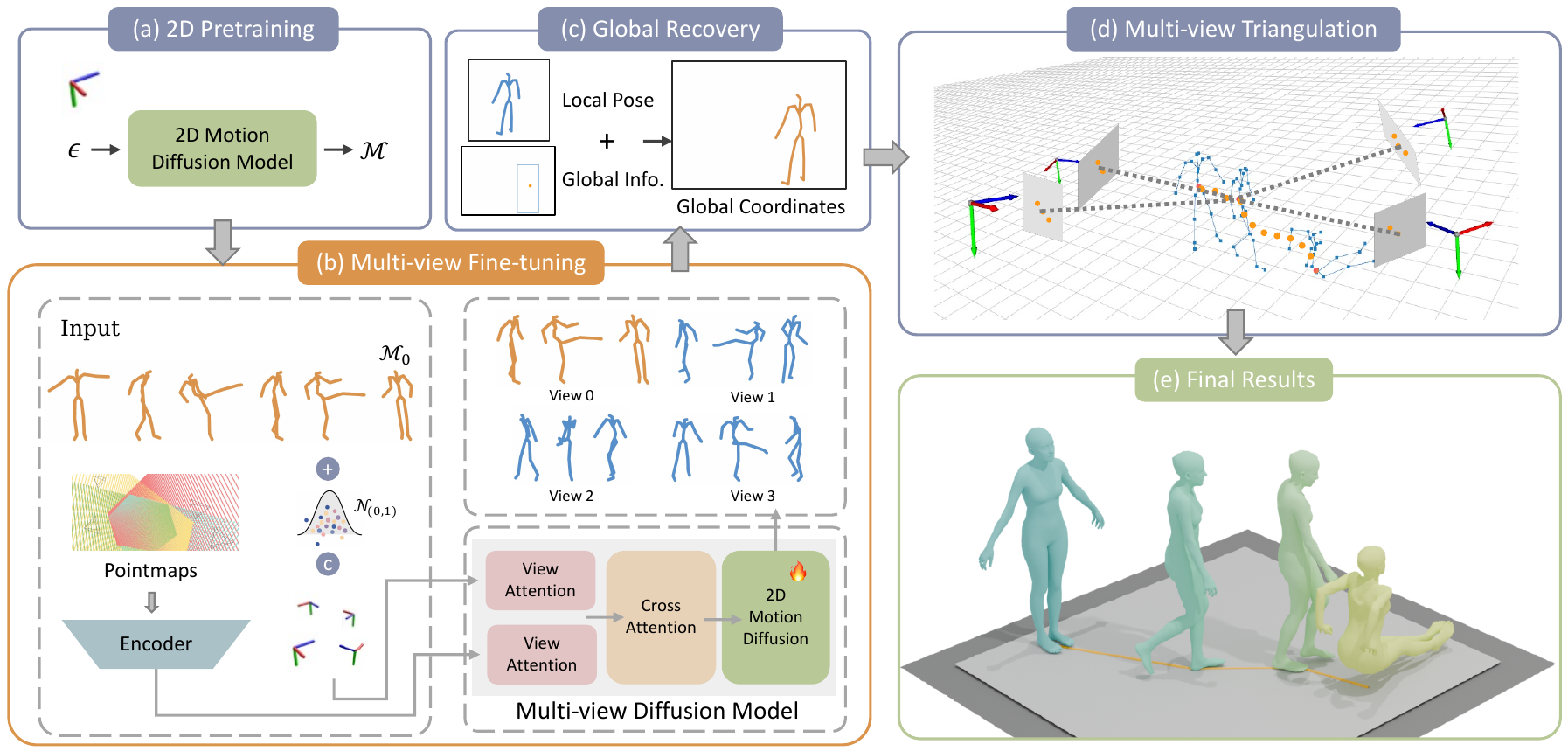}
\vspace{-6mm}
\caption{\textbf{Pipeline overview.}
     During training: (a) We first train an arbitrary single-view 2D Motion Diffusion Model. (b) Its weights are then used to initialize a Multi-view Diffusion Model, conditioned on 2D pose sequences from $\mathcal{V}_{0}$ and pointmaps. During inference, the Multi-view Model generates motions for other views. (c) We compute local poses and global movement to recover global coordinates for each view. (d) Multi-view triangulation is then used to synthesize 3D absolute poses, (e) resulting in full-body global human motion.
}
\label{fig:pipeline}
\vspace{-5mm} %3
\end{figure*}

We propose \textbf{Mocap-2-to-3}, a markerless motion capture multi-view lifting framework that lifts 2D poses to globally consistent 3D motion from monocular 2D sequences, as shown in \cref{fig:pipeline}.
We first pre-train a single-view 2D Motion Diffusion Model using 2D data (\cref{sec:pretrain}), then fine-tune a Multi-view Diffusion Model with public 3D data for multi-view consistency (\cref{sec:finetune}).
To recover human absolute positions in the world coordinate system, we extend the learning of local poses by additionally estimating global trajectory and scale across multiple views (\cref{sec:decouple}). 
For enhanced capture of global information, we encode the ground-plane equation into the network to accelerate convergence (\cref{sec:pointmaps}). 
Finally, we describe the inference pipeline that lifts 2D pose inputs to 3D motion and recovers the absolute positions of the human body (\cref{sec:inference}).
Our method retains the monocular input but leverages a multi-view diffusion model that captures cross-view motion priors, enables more consistent and accurate 3D motion lifting.

\subsection{2D motion pretraining}
\label{sec:pretrain}
Existing methods for 3D motion estimation typically require 3D data as labels for training. We explore how to better generalize the model to out-of-distribution scenarios. To this end, we reformulate 3D motion as a multi-view synthesis process and divide the training into two stages: single-view 2D pretraining and multi-view fine-tuning. In the first stage, we train a 2D motion generator for arbitrary camera viewpoints, termed the 2D Motion Diffusion Model. 
This stage establishes a strong motion prior by leveraging diverse 2D data, including real-world or publicly available videos during pretraining.
% This pretraining strategy facilitates and enables the use of large-scale 2D data in subsequent stages.

% To train the 2D diffusion model, we incorporate two types of training data: 1) 
% Re-projected 2D joints from HumanML3D~\cite{guo2022generating} (using camera parameters), with each batch trained on a single random view; 2) 2D data sharing distributional similarity with the test set. 
Following \cite{2023Human,pi2024motion}, we employ a transformer-based~\cite{vaswani2017attention} diffusion model~\cite{ho2020denoising} to implement the 2D Motion Diffusion model $\mathcal{D}_{2D}$. 
Diffusion-based architectures excel at modeling complex feature distributions and produce diverse yet coherent samples across viewpoints, offering clear advantages over deterministic regression backbones.
The network takes random noise $\epsilon$ as input and outputs a 2D motion sequence $\mathcal{M} \in \mathbb{R} ^{T\times J\times 2} $, where $T$ represents the number of video frames, and $J$ is the keypoint count.
In this stage, the model learns geometric relations to generate 2D motions from arbitrary camera views. By learning single-view generation, the pre-trained model can leverage large-scale 2D data to acquire motion priors across diverse viewpoints, which in turn accelerates convergence during fine-tuning.

\subsection{Multi-view fine-tuning}
\label{sec:finetune}
With the motion prior established, the standalone 2D Motion Diffusion model cannot ensure geometric consistency across views. Therefore, in the second stage, we fine-tune the 2D Motion Diffusion model with multi-view 2D supervision derived from 3D motion to enforce cross-view consistency for coherent 3D motion reconstruction. This stage further enables the model to learn canonical representations of human motion.

During fine-tuning, the number of viewpoints $V$ is set to 4, including a primary camera $\mathcal{V}_{0}$ (used for inference) and three virtual cameras whose poses are randomly sampled from the camera poses seen during the pretraining stage.
% For testing, the additional virtual viewpoints are chosen from test set camera perspectives. In practice, arbitrary viewpoints from the 2D pretraining stage can be incorporated to construct a virtual multi-view system.
Camera configuration details are provided in the supplementary material. 
To train the Multi-view Diffusion model $\mathcal{D}_{mv}$, we project 3D motion into each camera view to obtain geometrically consistent 2D motion ground truth. As no image pairs are required as input, we can apply random augmentations to existing 3D motion, including rotation, translation, and camera viewpoint augmentation (e.g., modifying pitch, yaw, roll, and distance). This enables large-scale virtual data generation from limited samples, enhancing model generalization.

The multi-view generation model~\cite{kapon2024mas} independently generates motions across views but lacks explicit consistency constraints. In our framework, we initialize $\mathcal{D}_{2D}$ with pre-trained weights and incorporate View Attention layers to enforce multi-view consistency. 
For \textbf{motion capture}, the model accepts Gaussian noise $\epsilon$, the primary view's 2D motion embedding $\mathcal{M}_{0} \in \mathbb{R} ^{T\times J\times 2} $, and camera embeddings (including camera intrinsics $\mathcal{K} \in \mathbb{R} ^{V\times 4} $ and extrinsics $\mathcal{RT}  \in \mathbb{R} ^{V\times 3} $) as its input.
 Geometrically consistent virtual-view 2D motions (see Sec.~3.3 for motion representation details) are generated from $\mathcal{M}_{0}$ and subsequently triangulated into 3D motions.
 % This design enhances cross-view coherence and improves generalization to unseen viewpoints during 3D motion reconstruction.
% For motion capture, the model takes both noise and the 2D motion embedding $\mathcal{M}_{0} \in \mathbb{R} ^{T\times J\times 2} $ from $V_{0}$ as input.
% \input{figs/decouple}

\subsection{Decomposed Motion Representation}
\label{sec:decouple}
% With our multi-view motion generation framework established, we now address the core task: recovering absolute human poses in world coordinates. The projection from 3D to 2D motion is as follows: any 3D pose in world coordinates can be mapped to 2D image global coordinates $\mathcal{M}^{g} \in \mathbb{R} ^{T\times J\times 2} $ via perspective projection using camera intrinsics and extrinsics. The global coordinates $\left ( u_{t,j}, v_{t,j} \right )$ of each frame keypoint represent its location in the original image, as shown in \cref{fig:decouple}(a).
% Multi-view 3D reconstruction is the inverse process: it first estimates 2D keypoints $\left ( u, v \right )$ in each view, then reconstructs global motion in world coordinates via triangulation~\cite{wiki_triangulation}.
% \textcolor{red}{The key challenge lies in estimating global coordinates of other views from a monocular input.}

\begin{figure}[ht]
\centering%
\includegraphics[width=\linewidth]{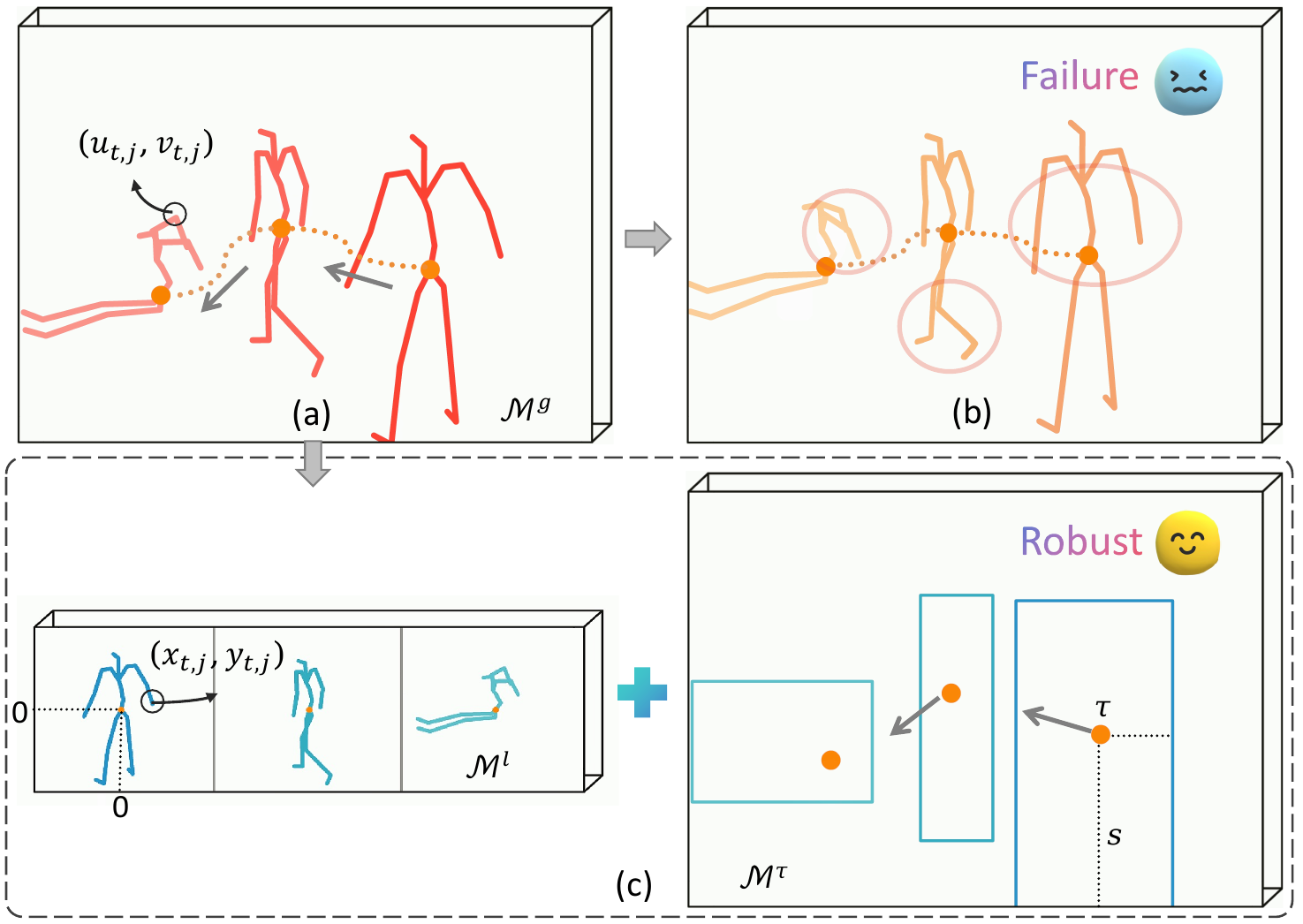}
\vspace{-7mm}
\caption{
%Decoupled representation for local pose and global movement (Refer to Sec. 3.3 for details).
(a) 2D projection coordinates, (b) direct prediction results (failure case). (c) Our decoupled representation separating local pose and global movement.
}
\label{fig:decouple}
\vspace{-3mm}
\end{figure}

% Since position has a much greater influence on the loss than skeletal structure, directly predicting $\left ( u_{t,j}, v_{t,j} \right )$ causes the network to focus more on positional cues than motion, leading to poor performance in motion details (\cref{fig:decouple}(b)). 
With the multi-view motion generation framework established, we turn to the other core challenge of recovering the absolute 3D position of motions in the world coordinate. To support the reconstruction from 2D poses to absolute 3D poses, We represent each 2D motion as the projection of a 3D motion in global coordinates under a specific camera viewpoint (\cref{fig:decouple}(a)). 
A straightforward approach would be to directly predict projected global coordinates from the given view; however, since position has a much greater influence on the loss than skeletal structure, such prediction tends to make the network focus more on positional cues rather than motion, resulting in degraded motion detail reconstruction (\cref{fig:decouple}(b)). The main challenge is to achieve globally consistent yet detail-preserving motion reconstruction.
To address this, we propose a novel human motion representation that decouples local pose and global movement, enabling independent optimization of action and trajectory. As shown in \cref{fig:decouple}(c), the local pose $\mathcal{M}^{l} \in \mathbb{R} ^{T\times (J-1)\times 2} $ without root position is obtained by cropping the 2D pose within bounding boxes, normalizing it to $\left [ -1,1 \right ] $, and centering the root joint to \textbf{remove root position influence}. The $j$-th keypoint is represented as $\left ( x_{t,j}, y_{t,j} \right )$. 
% The global movement $\mathcal{M} ^{\tau }  \in \mathbb{R} ^{T\times 2\times 2} $ includes the root trajectory $\tau =\left ( u^{\tau}_t, v^{\tau}_t   \right ) $ and motion scale $s =\left ( u^{s}_t, v^{s}_t   \right ) $, corresponding to the bounding box center and scale. 
The global movement $\mathcal{M}^{\tau} = [\tau, s] \in \mathbb{R}^{T\times2\times2}$ consists of the root trajectory $\tau \in \mathbb{R}^{T\times2}$ and the motion scale $s \in \mathbb{R}^{T\times2}$, corresponding to the pixel coordinates of the bounding box center and scale along the horizontal and vertical axes, respectively.
Our multi-view model predicts $\mathcal{M}_{v} \in \mathbb{R} ^{V\times T\times (J+1)\times 2} $, comprising (1) root-centered local poses $\mathcal{M}_{v}^{l}$ (a (J-1)-dimensional vector), and (2) global movement $\mathcal{M}_{v}^{\tau}$ (a 2-dimensional vector). 
% To improve $\mathcal{M}_{v}^{\tau}$ estimation, we condition on both camera intrinsics ($\mathcal{K} \in \mathbb{R} ^{V\times 4} $) and extrinsics ($\mathcal{RT}  \in \mathbb{R} ^{V\times 3} $) as input.

During inference, given a monocular input, the model generates virtual-view outputs $\mathcal{M}_{v}^{l}$ and $\mathcal{M}_{v}^{\tau}$ for each additional viewpoint. The transformation from multi-view local to global coordinates $\mathcal{M}_{v}^{g}  \in \mathbb{R} ^{V\times T\times J\times 2} $ is then computed as follows:

\begin{equation}
{\footnotesize  
\begin{aligned}
    \mathcal{M}_{v ,\left \{ 1:J \right \}  }^{g} &= \mathcal{M}_{v}^{l} \cdot s_v + \tau_v, \\
    \mathcal{M}_{v}^{g} &= [\tau_v\ , \mathcal{M}_{v, \left \{ 1:J \right \} }^{g}].
\end{aligned}
}
\label{eq:global}
\end{equation}

Here, $\mathcal{M}_{v}^{l}$ is used to compute the global coordinates of all joints except the root using $s_v$ and $\tau_v$, and then concatenated with the root coordinate $\tau_v$. Finally, the multi-view $\mathcal{M}_{v}^{g}$ is used to reconstruct absolute 3D poses through camera parameters and triangulation~\cite{wiki_triangulation}.

\subsection{Ground Constraint Encoding}
\label{sec:pointmaps}

\begin{figure}[ht]
\centering%
\includegraphics[width=0.7\linewidth]{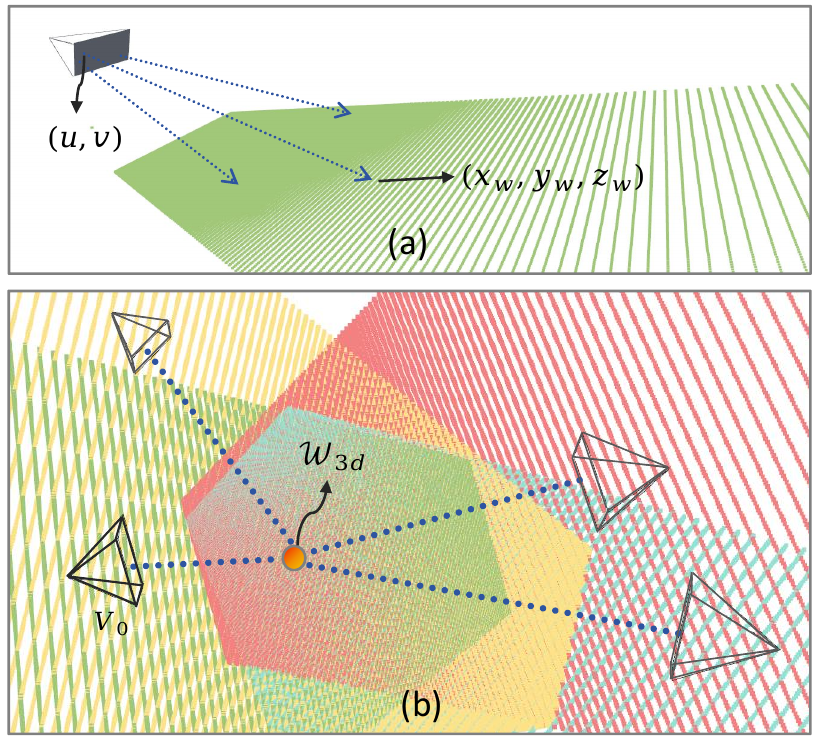}
\vspace{-2mm}
\caption{(a) Pointmaps representing pixel-to-world coordinate $(u,v)\leftrightarrow (x_w,y_w,z_w)$ mappings. (b) Multi-view pointmaps in world coordinate system.
}
\vspace{-4mm}
\label{fig:pointmap}
\end{figure}

% \textcolor{red}{Accurate human localization is crucial for interaction with the physical world.}
In monocular-to-multi-view generation, pose learning is easier due to normalized representations across viewpoints, while movement scales vary significantly. However, due to depth ambiguity in monocular settings, learning 2D motion locations for other views from a source view $\mathcal{V}_{0}$ converges slowly, even when camera embeddings are provided as input. 
Our idea is to leverage physical-world constraints to enhance both localization accuracy and efficiency. To this end, we design a geometric encoding scheme: we introduce explicit geometric constraints by leveraging known camera poses to compute ground planes, which are then represented as intuitive pointmaps~\cite{leroy2024grounding,wang2024dust3r}. These constraints substantially accelerate the convergence of network training for position learning.

In our work, pointmaps $\mathcal{P} \in \mathbb{R} ^{W\times H\times 3} $ represent the mapping from each pixel $\left ( u,v \right )$ in an image $I$ of resolution $W\times H$ to its corresponding 3D point $\left ( x_{w},y_{w},z_{w} \right )$ in world coordinates, as shown in \cref{fig:pointmap}(a), i.e.,$I_{u,v} \leftrightarrow P_{x_{w},y_{w},z_{w}}$. For any dataset with a world coordinate system grounded on the ground plane, this mapping can be directly computed given the camera intrinsics and extrinsics. Each point is the intersection of a ray from the camera center with the ground plane, forming a view-specific ground point cloud. The detailed computation is provided in the supplementary material. It is important to note that we only include the ground plane rather than full environmental point clouds, as pointmaps can be \textbf{computed directly from camera poses without additional sensors or ground-truth scans}. This avoids extra cost and facilitates real-world deployment. Representing the ground plane as pointmaps allows the network to learn more intuitively, providing a natural 2D–to-3D correspondence across views (\cref{fig:pointmap}(b)).

As shown in \cref{fig:pipeline}(b), pointmaps are incorporated as a conditioning input, first compressed into feature representations through a ResNet-18\cite{he2016deep} encoder, and then integrated into $\mathcal{D}_{mv}$ with two attention layers: a View Attention Layer to learn cross-view correlations and a Cross Attention Layer to guide the generation of motion $\mathcal{M}_v$. Pointmaps accelerate the convergence of global movement learning and serve as a plug-and-play module for introducing explicit geometric constraints in any multi-view global estimation task.

\subsection{Inference}
\label{sec:inference}

During inference, the denoising process comprises $N$ steps. For each timestep $n$, model $\mathcal{D}_{mv}$ takes
$\left [ \epsilon ,\mathcal{M} _{0},\mathcal{K},\mathcal{RT},\mathcal{P} \right ] $ as input and predicts the 2D motion sequence $\mathcal{M}_{v}^{n}$. For any viewpoint, $\mathcal{M}_{v}^{n}$ is transformed to $\mathcal{M} _{v}^{gn}$ using \cref{eq:global}, and $\mathcal{M} _{v}^{gn}$ is triangulated~\cite{wiki_triangulation} to obtain the 3D absolute pose $\mathcal{W} _{3d}^{n} \in \mathbb{R} ^{T\times J\times 3} $ in the world coordinate system. To enforce multi-view consistency, we project $\mathcal{M} _{v}^{gn}$ into each view to recompute $\mathcal{M} _{v}^{ln}$ and $\mathcal{M} _{v}^{\tau n}$, then update $\mathcal{M} _{v}^{n-1}$ from the previous step. At the final timestep $N$, we obtain the 3D motion $\mathcal{W} _{3d}^{0}$ with global position. Following \cite{pi2024motion,kapon2024mas}, if SMPL~\cite{15tog_smpl} parameters are required on top of the reconstructed 3D poses, SMPLify~\cite{bogo2016keep} can be applied as a post-hoc fitting step to estimate the parameters from the recovered joints. Refer to the supplementary material for pseudocode.
% we apply SMPLify~\cite{bogo2016keep} to fit SMPL~\cite{15tog_smpl} pose parameters.

\section{Experiment}
\begin{table*}[t!]
\small
\centering
\setlength{\tabcolsep}{5pt}
\renewcommand{\arraystretch}{0.85}
\begin{tabular}{lcccccccc}
\toprule
& \multicolumn{2}{c}{Camera Coordinate} & \multicolumn{2}{c}{World Alignment} & \multicolumn{1}{c}{World Coordinate} & \multicolumn{3}{c}{Motion Quality} \\
\cmidrule(lr){2-3}\cmidrule(lr){4-5}\cmidrule(lr){6-6}\cmidrule(lr){7-9}
Method & PA-MPJPE$\downarrow$ & MPJPE$\downarrow$ & W-MPJPE$\downarrow$ & WA-MPJPE$\downarrow$ & Abs-MPJPE$\downarrow$ & Accel$\downarrow$ & Jitter$\downarrow$ & FS$\downarrow$ \\
\midrule
SMPLify$^*$~\cite{bogo2016keep} & 83.8 & 155.3 & 284.4 & 165.7 & 406.2 & 28.6 & 371.5 & 57.9 \\
\midrule
MotionBERT~\cite{zhu2023motionbert} & 123.0 & 146.0 & -- & -- & -- & -- & -- & -- \\
\midrule
 SA-HMR$^*$$^\ddagger$$^\dagger$~\cite{shen2023learning} & 51.1 & 93.2 & -- & -- & \cellsecond{268.3} & -- & -- & -- \\
\midrule
WHAM$^*$~\cite{shin2024wham} & 40.1 & 74.4 & 182.5 & 106.1 & -- & 4.9 & 17.4 & \cellsecond{3.5} \\
GVHMR$^*$~\cite{shen2024world} & 33.6 & 58.9 & 110.0 & \cellsecond{68.4} & -- & 3.8 & 10.5 & \cellfirst{\textbf{2.5}} \\
TRAM$^*$$^\dagger$~\cite{wang2024tram} & 36.3 & 67.1 & 169.3 & 107.9 & 533.8 & 4.3 & 173.2 & 27.6 \\
\midrule
 WHAM+SMPLify$^*$$^\dagger$~\cite{shin2024wham,bogo2016keep} & 37.5 & 68.9 & 173.2 & 101.3 & 456.3 & 5.4 & 14.4 & 5.7 \\
 GVHMR+SMPLify$^*$$^\dagger$~\cite{shen2024world,bogo2016keep} & \cellsecond{30.7} & \cellsecond{58.7} & \cellsecond{109.4} & 68.6 & 430.4 & \cellsecond{3.7} & \cellsecond{9.4} & 5.6 \\
\midrule
 \textbf{Ours$^\dagger$} & \cellfirst{\textbf{26.2}} & \cellfirst{\textbf{39.6}} & \cellfirst{\textbf{82.6}} & \cellfirst{\textbf{50.1}} & \cellfirst{\textbf{156.8}} & \cellfirst{\textbf{2.5}} & \cellfirst{\textbf{8.0}} & \cellsecond{3.5} \\
\bottomrule
\end{tabular}
\vspace{-2mm}
\caption{\textbf{Quantitative results on RICH in:} (1) Root-aligned in Camera Coordinates, (2) World Coordinates with initial-frame alignment, (3) World Coordinates without any alignment. The symbols $^*$, $^\ddagger$, and $^\dagger$ denote the inclusion of images, scene scans, and calibrated camera poses as inputs, respectively. The best and second-best results are highlighted \textfirst{\textbf{green}} and \secondtext{yellow}.}
\label{tab:rich_comp}
\vspace{-3mm}
\end{table*}

\subsection{Datasets and Metrics}

\noindent \textbf{Training datasets.}
To pre-train the 2D diffusion model, we use two types of data: (1) projected 2D joints from HumanML3D~\cite{guo2022generating}, training each batch on a single random view; and (2) 2D data from the same source as the test set(e.g., the RICH training set). We then fine-tune the multi-view diffusion model on HumanML3D~\cite{guo2022generating}, BEDLAM~\cite{black2023bedlam}, and Human3.6M~\cite{ionescu2013human3}, where HumanML3D includes HumanAct12~\cite{guo2020action2motion} and AMASS~\cite{mahmood2019amass}.

\noindent \textbf{Evaluation datasets.}
Following \cite{shen2024world,li2025lifting}, we evaluate our model on RICH~\cite{huang2022capturing} and AIST++~\cite{li2021ai}, two real-world datasets covering both outdoor and indoor scenes. They include actions like sitting, lying down, and handstands, which are less represented in the training set and offer a more comprehensive test of the model’s generalization.

\noindent \textbf{Metrics.}
We follow the evaluation protocol~\cite{shin2024wham,shen2024world} and use standard metrics. In the camera coordinate system, we compute per-frame root-aligned Mean Per-Joint Position Error (MPJPE) and Procrustes-Aligned MPJPE (PA-MPJPE) to evaluate pose accuracy. For world coordinates, we use W-MPJPE (aligned to the first two frames) and WA-MPJPE (with full-sequence alignment) to assess global trajectories. Since our method predicts absolute world positions, we also compute \textbf{Abs-MPJPE (without any alignment)}. All position errors are reported in millimeters (mm). Additionally, we evaluate root translation error ($T_{root}$), motion smoothness (Accel/Jitter), and foot sliding (FS).

\begin{figure*}[ht]
\centering%
\includegraphics[width=\linewidth]{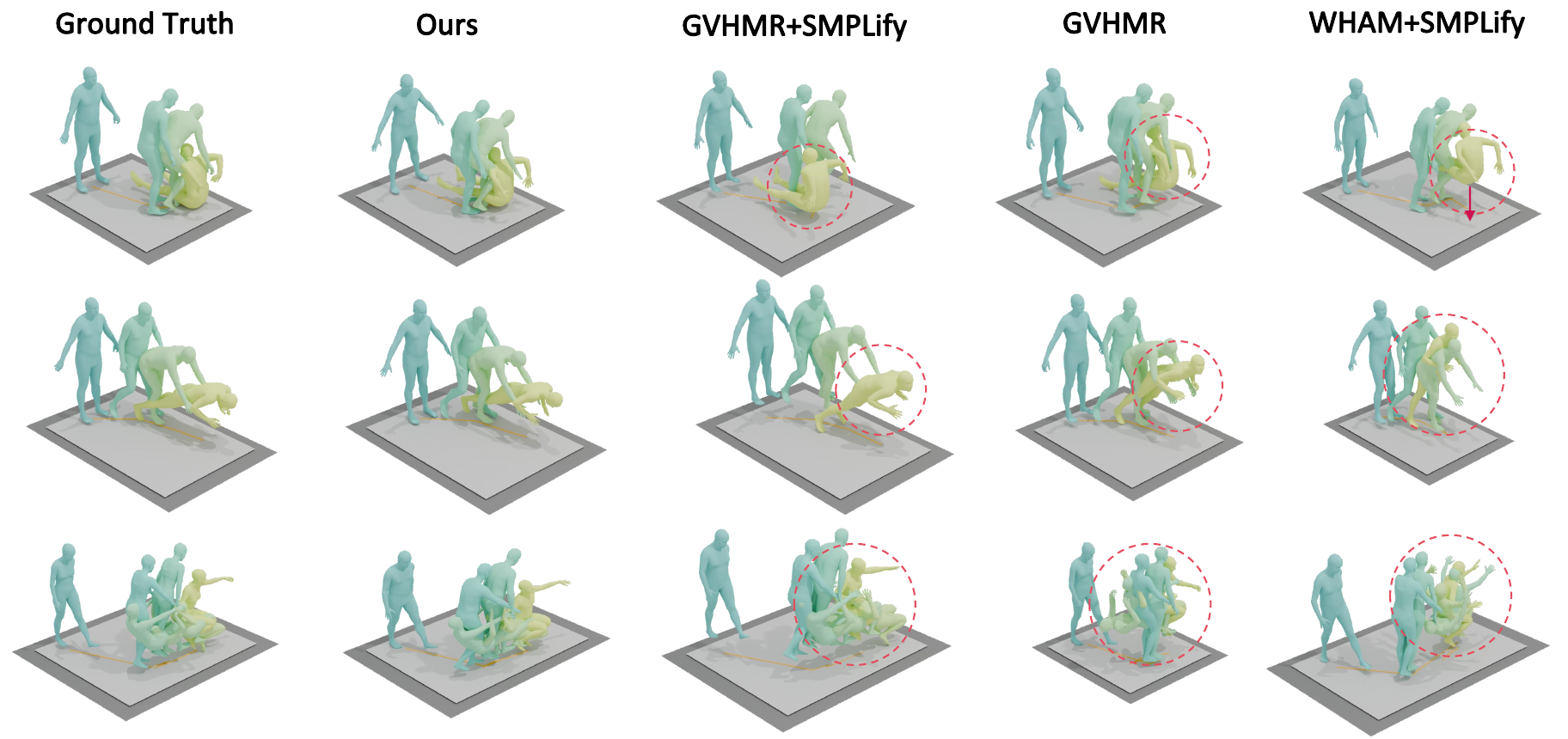}
\vspace{-7mm}
\caption{\textbf{Qualitative comparison on RICH.}
    % Compare global motions after first-frame alignment in world coordinates. Our method generates more realistic OOD motions with accurate body orientation and positioning. 
    Global motions are compared after first-frame alignment. Our method generates more realistic OOD motions with accurate body orientation and positioning, while red circles mark unnatural baseline poses.
    %Unnatural poses in baseline results are highlighted in red circles.
}
\vspace{-5mm}
\label{fig:comp}
\end{figure*}
\begin{figure*}[ht]
\centering%
\includegraphics[width=\linewidth]{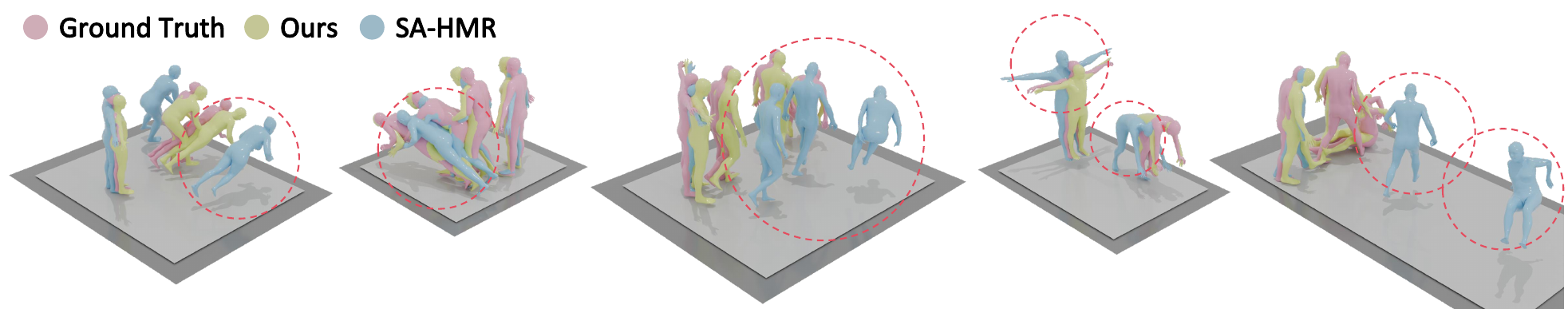}
\vspace{-8mm}
\caption{\textbf{Qualitative comparison on RICH.}
    %Compare unaligned absolute poses in the shared world coordinate system.
    Unaligned absolute pose comparison in shared world coordinates. Unlike baseline methods that exhibit positional drift, our solution maintains accurate localization without requiring additional equipment.
}
\vspace{-6mm}
\label{fig:comp_world}
\end{figure*}

\subsection{Lifting SMPL keypoints with ground truth}
Our method performs 2D-to-3D lifting and can process different keypoint formats. In this section, we analyze the most widely-used SMPL~\cite{15tog_smpl} format. Existing 3D motion prediction methods typically involve two stages: (1) detecting 2D keypoints~\cite{xu2022vitpose,sun2019deep}, and (2) predicting 3D poses from 2D keypoints~\cite{zhu2023motionbert,xu2024finepose}. We focus on the second stage: estimating 3D motion from 2D keypoints, and use ground-truth 2D keypoints for intuitive performance comparison, replacing all baselines' 2D keypoint inputs with ground truth for fairer comparison.

We compare our method against five categories of baselines:
(1) optimization-based methods represented by SMPLify~\cite{bogo2016keep};
(2) 2D-to-3D lifting models exemplified by MotionBERT~\cite{zhu2023motionbert};
(3) environment-aware models that predict absolute world-coordinate poses, such as SA-HMR~\cite{shen2023learning}; and
(4) state-of-the-art 3D data-dependent methods, including WHAM~\cite{shin2024wham}, GVHMR~\cite{shen2024world}, and TRAM~\cite{wang2024tram}.
(5) To ensure a fair comparison under identical input settings, we also evaluate two strong hybrid baselines: WHAM+SMPLify~\cite{shin2024wham,bogo2016keep} and GVHMR+SMPLify~\cite{shen2024world,bogo2016keep}.
For these hybrid baselines, we initialize SMPLify~\cite{bogo2016keep} with the outputs of WHAM~\cite{shin2024wham} or GVHMR~\cite{shen2024world}, and perform optimization using ground-truth 2D keypoints and calibrated camera poses.
This integration of state-of-the-art learning-based methods with optimization-based refinement represents the current best-performing paradigm for accurate 3D human motion recovery.

% It should be emphasized that although we compare with WHAM~\cite{shin2024wham} and GVHMR~\cite{shen2024world}, we address fundamentally different problems: \textbf{our primary objective is to recover absolute human poses in world coordinates from monocular input, rather than estimating root aligned global trajectories}, as this enables wider applications in interactive scenarios. We include comparisons with WHAM~\cite{shin2024wham} and GVHMR~\cite{shen2024world} to demonstrate that our method achieves comparable motion performance while additionally recovering absolute positioning. 
% Furthermore, to ensure fair comparison under identical input settings, we provide enhanced baselines by combining WHAM+SMPLify~\cite{shin2024wham,bogo2016keep} and GVHMR+SMPLify~\cite{shen2024world,bogo2016keep}, where we initialize SMPLify~\cite{bogo2016keep} with outputs from WHAM~\cite{shin2024wham} or GVHMR~\cite{shen2024world} and optimize using ground-truth 2D keypoints and actual camera poses, this integration of state-of-the-art methods with optimization represents the current best-performing approach in the field.

The quantitative results on RICH~\cite{huang2022capturing} are shown in \cref{tab:rich_comp}. The symbols $^*$, $^\ddagger$, and $^\dagger$ denote the inclusion of images, scene scans, and calibrated camera poses as inputs, respectively. In the camera coordinate system, PA-MPJPE and MPJPE assess motion quality after root alignment, removing positional effects. 
Compared to GVHMR+SMPLify \cite{shen2024world,bogo2016keep}, our method reduces PA-MPJPE by 4.5 mm, demonstrating stronger expressive power in reconstructing motion details.
% Although the regression+optimization baseline mainly serves to align inputs for fair comparison, its two-stage nature is less practical than our end-to-end approach. 
In world coordinate evaluation with temporal alignment, our method achieves superior global trajectory estimation. 
Furthermore, when compared with methods that also take calibrated camera poses as input (marked with $^\dagger$), our approach achieves lower errors in Abs-MPJPE, indicating more accurate estimation of absolute positions. Notably, this is achieved without requiring scanned environmental parameters as in SA-HMR~\cite{shen2023learning}, making our spatial constraints both more practical and easier to deploy in real-world settings.
Additionally, our results show smoother motion, and the slightly higher foot-sliding error compared to GVHMR~\cite{shen2024world} is due to our not using foot-sliding as a post-processing optimization like GVHMR~\cite{shen2024world}, which we plan to add in future.

Qualitative comparisons in \cref{fig:comp} show global trajectories after first-frame alignment. For out-of-distribution actions like squatting and bending, Our method generates more realistic poses. Regression-based methods, limited by their 3D training data, often fail to recover reasonable poses when initial estimates are poor, despite 2D keypoint optimization. In contrast, our pretraining effectively learns motion priors that improve generalization to unseen actions. \cref{fig:comp_world} compares absolute poses in world coordinates (without alignment) between our method and SA-HMR~\cite{shen2023learning} in a shared coordinate system, highlighting global positioning differences. SA-HMR~\cite{shen2023learning} shows notable errors in global position and body scale, while our results align more closely with ground truth.

\subsection{Lifting COCO keypoints with detector}

To demonstrate the effectiveness of different keypoint formats, we trained a COCO version of the lifting model and present results using the 2D detector ViTPose~\cite{xu2022vitpose} as input. We selected AIST++\cite{li2021ai} as the test set, which provides 3D ground truth in COCO format. Following\cite{li2025lifting}, we compare our method with baselines including ElePose~\cite{wandt2022elepose}, MAS~\cite{kapon2024mas}, SMPLify~\cite{bogo2016keep}, MotionBERT~\cite{zhu2023motionbert}, WHAM~\cite{shin2024wham}, GVHMR~\cite{shen2024world}, and MVLift~\cite{li2025lifting}.

\begin{table}[tbp]
    \centering
    \footnotesize
    \renewcommand{\arraystretch}{0.85}
    \begin{tabular}{lccc}
        \toprule
        Methods & PA-MPJPE$\downarrow$ & MPJPE$\downarrow$ & $T_{\text{root}}$$\downarrow$ \\
        \midrule
        ElePose~\cite{wandt2022elepose}   & 251.1 & 269.4 & -- \\
        MAS~\cite{kapon2024mas}       & 155.6 & 191.1 & -- \\
        SMPLify$^*$~\cite{bogo2016keep}   & 146.7 & 171.6 & 77.4 \\
        \midrule
        MotionBERT~\cite{zhu2023motionbert} & 108.6 & 134.0 & 101.6 \\
        WHAM$^*$~\cite{shin2024wham}       & 75.1  & 104.8 & 164.3 \\
        GVHMR$^*$~\cite{shen2024world}     &  68.6 & 119.7 & 253.1 \\
        GVHMR+SMPLify$^*$$^\dagger$~\cite{shen2024world,bogo2016keep}  & \cellsecond{62.2} & \cellsecond{102.8} & 112.3 \\
        \midrule
        MVLift~\cite{li2025lifting}     & 79.2 & 110.7 & \cellsecond{67.6} \\
        \textbf{Ours$^\dagger$}      & \cellfirst{\textbf{60.1}} & \cellfirst{\textbf{90.9}} & \cellfirst{\textbf{61.8}} \\
        \bottomrule
    \end{tabular}
    \vspace{-2mm}
    \caption{\textbf{Quantitative results on AIST++.} Symbols $^*$ or $^\dagger$ indicate the use of images or calibrated camera poses as inputs.}
    \label{tab:aist_comp}
    \vspace{-4mm}
\end{table}

\begin{figure}[ht]
\centering%
\includegraphics[width=\linewidth]{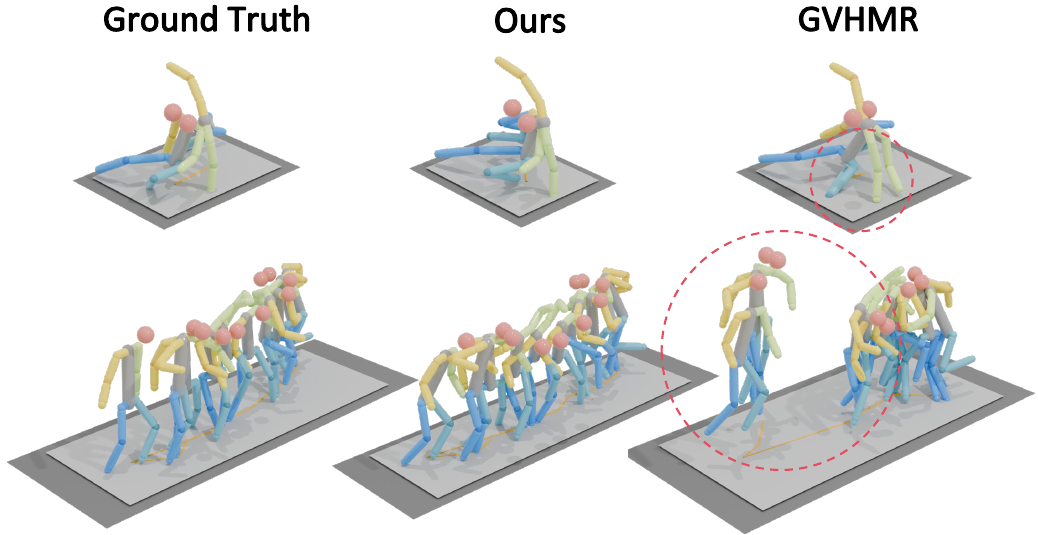}
\vspace{-7mm}
\caption{\textbf{Qualitative comparison on AIST++.} Our method generalizes well to COCO-format skeletons as well.
}
\vspace{-5mm}
\label{fig:aist_comp}
\end{figure}

Quantitative results are presented in \cref{tab:aist_comp}. Both MVLift~\cite{li2025lifting} and our method are 2D-to-3D reconstruction approaches. 
Our method outperforms MVLift~\cite{li2025lifting} and GVHMR+SMPLify\cite{shen2024world,bogo2016keep} in both motion accuracy evaluation (PA-MPJPE) and global trajectory assessment ($T_{root}$).
Qualitatively, as demonstrated in \cref{fig:aist_comp}, our approach maintains robust performance even for challenging dance motions, with neither global trajectory nor foot-ground contact exhibiting unrealistic deviations.

\subsection{Ablation study}
\label{sec:abs}
We conduct ablation studies of Mocap-2-to-3 on the RICH~\cite{huang2022capturing} dataset, with results in \cref{tab:ablation}. Comparing rows 1–2 shows that decoupling local pose and global movement representations significantly improves action recognition and trajectory estimation. 
Rows 2–4 show that without pointmaps, convergence is slower at the same epoch. While pointmaps are not essential for model learning, training to 8k epochs yields comparable performance, showing that pointmaps reduce training time by over 50\%.
% The final rows show that using the RICH 2D training set in pretraining substantially improves motion quality, as reflected in PA-MPJPE and MPJPE. 
The final rows show that adding just 175 in-domain RICH 2D sequences during pretraining substantially improves motion quality, as reflected in PA-MPJPE and MPJPE.
Even without such data, our method outperforms GVHMR+SMPLify\cite{shen2024world,bogo2016keep}, confirming the robustness of our lifting framework.
These findings suggest that simply using 2D data can effectively enhance 3D motion estimation performance, opening new possibilities for improving the generalization capability of human 3D motion recovery models.

\begin{table}[tbp]
    \centering
    \resizebox{1.0\linewidth}{!}{
    \setlength{\tabcolsep}{1mm}
    \renewcommand{\arraystretch}{0.85}
    \begin{tabular}{lccccc}
        \toprule
        Methods & PA-MPJPE$\downarrow$ & MPJPE$\downarrow$ & Abs-MPJPE$\downarrow$ & W-MPJPE$\downarrow$ & Epoch \\
        \midrule
        w/o decouple     & 65.1 & 121.3 & 544.2 & 161.2 & -- \\
        w/o pointmaps    & 45.8 & 85.6  & 373.9 & 121.8 & 3.5k \\
        w/o pointmaps    & 33.4 & 52.3  & 182.5 & 103.7 & 8k \\
        w/ pointmaps     & \cellsecond{30.5} & \cellsecond{45.3} & \cellsecond{157.9} & \cellsecond{88.6} & 3.5k \\
        \textbf{w/ 2D RICH}       & \cellfirst{\textbf{26.2}} & \cellfirst{\textbf{39.6}} & 
                           \cellfirst{\textbf{156.8}} & \cellfirst{\textbf{82.6}} & 3.5k \\
        \bottomrule
    \end{tabular}
    }
    \vspace{-2mm}
    \caption{\textbf{Ablation study on RICH:} Pointmaps boost convergence; 2D pretraining increases motion accuracy.}
    \label{tab:ablation}
    \vspace{-5mm}
\end{table}

\section{Limitation and Future work}
\label{sec:future}

While our approach demonstrates competitive performance, certain limitations remain. Our framework is capable of lifting arbitrary-format 2D skeletons to high-quality 3D motion. However, we observe that inaccurate 2D skeletons obtained from raw videos can degrade the quality of the reconstructed 3D motion. Importantly, this is not a limitation of our framework itself: when provided with more reliable 2D poses (e.g., manually annotated SMPL-format poses or ViTPose~\cite{xu2022vitpose}-generated COCO-format poses), our method works effectively and maintains strong performance. In future work, we aim to further improve the 2D SMPL-format motion prediction component and explore incorporating detection confidence during training to enhance robustness.
We also plan to integrate additional geometric constraints, such as foot-sliding reduction, to further enhance model fidelity.
Furthermore, we are exploring interaction tasks built upon this work, targeting embodied and gaming scenarios, aiming to further validate the method’s potential in complex real-world applications.

% Furthermore, we plan to incorporate post-processing for foot sliding reduction and to explore human-human and human-scene interaction, with the goal of further validating the model's performance in complex, real-world scenarios.}

% Furthermore, building upon this foundation of recovering absolute poses in the physical world, we will focus future work on human-human and human-scene interaction, to further validate the model's performance.}
% Furthermore, we plan to incorporate post-processing for foot sliding reduction and replace camera calibration with SLAM-based camera pose estimation to further enhance the flexibility and fidelity of the model.

\section{Conclusion}
\label{sec:concl}
We present \textbf{Mocap-2-to-3}, a novel framework that leverages accessible 2D data to enhance 3D estimation and recover metrically accurate poses from monocular input.
Overall, our approach \textbf{bridges the gap} between conventional human motion recovery and markerless motion capture, enabling accurate and physically consistent motion estimation in real-world scenes, and offering a promising direction for future research in this domain.
% Experimental results further demonstrate its ability to achieve more accurate absolute positioning and improve 3D motion estimation in out-of-distribution scenarios using only supplementary 2D data, offering a promising direction for future research in this domain.

% The key contribution is that we leverages homologous 2D data to enhance 3D motion estimation, especially in out-of-distribution scenarios. 
% To recover more fine-grained motions and positions in the physical world, we separately learn local poses and global movements, and design a geometric constraint encoding to obtain more accurate global locations.

% \textcolor{red}{Since our method estimates absolute human positions, it is readily applicable to both single-person and multi-person interaction scenarios, simply via multiple inference passes.}
% Experimental results demonstrate the potential of our method to improve 3D motion estimation in out-of-distribution scenarios using only supplementary 2D data, offering a promising direction for future research in this domain.

\noindent\textbf{Acknowledgments.} This work is supported in part by the Opening Project of the State Key Laboratory of General Artificial Intelligence, BIGAI/Peking University, Beijing, China (Project No.~SKLAGI2025OP17), and Deep Glint.
{
    \small
    \bibliographystyle{ieeenat_fullname}
    \bibliography{main}
}


\end{document}